\documentclass[lettersize,journal]{IEEEtran}
\usepackage[driverfallback=dvipdfm,
CJKbookmarks=true, bookmarksnumbered=true, bookmarksopen=true,
colorlinks=true, %
pdfborder=001, %
citecolor=blue, linkcolor=blue, anchorcolor=red, urlcolor=black
]{hyperref}
\usepackage{caption}
\usepackage{amsmath,amsfonts}
\usepackage{algorithmic}
\usepackage{algorithm}
\usepackage{array}
\usepackage[caption=false,font=normalsize,labelfont=sf,textfont=sf]{subfig}
\usepackage{textcomp}
\usepackage{stfloats}
\usepackage{url}
\usepackage{verbatim}
\usepackage{graphicx}
\usepackage{color}
\usepackage{cite}
\usepackage{booktabs}
\usepackage{pifont}
\usepackage{multirow}
\usepackage{makecell}
\usepackage{amsfonts}
\usepackage{threeparttable}
\newcommand{\tablestyle}[2]{\setlength{\tabcolsep}{#1}\renewcommand{\arraystretch}{#2}\centering\small}
\newcommand{\cmark}{\ding{51}}
\newcommand{\xmark}{\ding{55}}
\usepackage[table]{xcolor}
\usepackage{arydshln}
\definecolor{grey}{RGB}{232,232,232}
\definecolor{pblue}{RGB}{224,250,250}
\definecolor{lightgrey}{RGB}{255,255,255}
\definecolor{baselinecolor}{gray}{.9}
\newlength\savewidth

\begin{document}

\title{MAR: \underline{M}asked Autoencoders for \\ Efficient \underline{A}ction \underline{R}ecognition}

\author{Zhiwu~Qing,
        Shiwei~Zhang,
        Ziyuan~Huang,
        Xiang~Wang, 
        Yuehuan~Wang,
        Yiliang~Lv,\\
        Changxin~Gao,
        Nong~Sang}

\markboth{Journal of \LaTeX\ Class Files,~Vol.~14, No.~8, August~2021}%
{Shell \MakeLowercase{\textit{et al.}}: A Sample Article Using IEEEtran.cls for IEEE Journals}


\maketitle
\let\thefootnote\relax\footnotetext{This work is supported by the National Natural Science Foundation of China under grant 61871435, Fundamental Research Funds for the Central Universities no.2019kfyXKJC024, by the 111 Project on Computational Intelligence and Intelligent Control under Grant B18024, and by Alibaba Group through Alibaba Research Intern Program. \textit{(Corresponding author: N. Sang and S. Zhang.)}}
\let\thefootnote\relax\footnotetext{Z. Qing, X. Wang, Y. Wang, C. Gao, and N. Sang are with the National Key Laboratory of Science and Technology on Multispectral Information Processing, School of Artificial Intelligence and Automation, Huazhong University of Science and Technology, Wuhan 430074, China (e-mail: qzw@hust.edu.cn, wxiang@hust.edu.cn, yuehwang@hust.edu.cn, cgao@hust.edu.cn and nsang@hust.edu.cn).}
\let\thefootnote\relax\footnotetext{S. Zhang and Y. Lv are with Alibaba Group, Hangzhou 311100, Zhejiang, China (e-mail: zhangjin.zsw@alibaba-inc.com and yiliang.lyl@alibaba-inc.com).}
\let\thefootnote\relax\footnotetext{Z. Huang is with Advanced Robotics Centre at National University of Singapore, Singapore (e-mail: ziyuan.huang@u.nus.edu)}

\begin{abstract}
Standard approaches for video action recognition usually operate on the full input videos, which is inefficient due to the widely present spatio-temporal redundancy in videos. 
Recent progress in masked video modelling, \textit{i.e.,} VideoMAE, has shown the ability of vanilla Vision Transformers (ViT) to complement spatio-temporal contexts given only limited visual contents. Inspired by this, we propose propose Masked Action Recognition (MAR), which reduces the redundant computation by discarding a proportion of patches and operating only on a part of the videos.
%
%
MAR contains the following two indispensable components: \textit{cell running masking} and \textit{bridging classifier}.
Specifically, to enable the ViT to perceive the details beyond the visible patches easily, cell running masking is presented to preserve the spatio-temporal correlations in videos, \textit{i.e.,} it ensures the patches at the same spatial location can be observed in turn for easy reconstructions. 
%
%
Additionally, we notice that, although the partially observed features can reconstruct semantically explicit invisible patches, they fail to achieve accurate classification.
To address this, a bridging classifier is proposed to bridge the semantic gap between the ViT encoded features for reconstruction and the features specialized for classification.
Our proposed MAR reduces the computational cost of ViT by 53\% and extensive experiments show that  MAR consistently outperforms existing ViT models with a notable margin. 
Especially, we found a ViT-Large trained by MAR outperforms the ViT-Huge trained by a standard training scheme by convincing margins on both Kinetics-400 and Something-Something v2 datasets, while our computation overhead of ViT-Large is only 14.5\% of ViT-Huge.
Codes and models will be made available \href{https://github.com/alibaba-mmai-research/Masked-Action-Recognition}{\textcolor{blue}{here}}.
\end{abstract}


\section{Introduction}

In recent years, deep neural networks~\cite{feichtenhofer2019slowfast,feichtenhofer2020x3d,wang2021tdn,liu2021videoswin,fan2021mvit,arnab2021vivit} have achieved impressive performance for the action recognition task on several large-scale video datasets~\cite{kay2017k400,goyal2017something}. 
%
%
These methods usually depend on full video frames to understand the visual content.
Although this yields decent performances, the computation over full videos is highly redundant due to the excessive and widely present spatio-temporal redundancy of visual information~\cite{tong2022videomaenju,wang2021adafocus,wang2021adafocusv2,wu2019adaframerel3,korbar2019scsamplerrel4} in videos.
%
%
%
%
In light of this, a branch of previous works has proposed to reduce the spatio-temporal redundancy by training an additional model to focus on relevant frames~\cite{wu2019rel1,wu2019liteevalrel2,wu2019adaframerel3,korbar2019scsamplerrel4,li20212dada3drel5,meng2020arrel6} or spatio-temporal regions~\cite{wang2021adafocus,wang2021adafocusv2}, which can significantly reduce the computation cost.
However, they mostly require complicated operations, such as reinforcement learning and multi-stage training.

%


In self-supervised video representation learning, Masked Autoencoders (MAE)~\cite{tong2022videomaenju,feichtenhofer2022videomaefb} discard a high proportion of vision patches to yield a non-trivial and meaningful self-supervised reconstruction task.
%
The simple masking strategy with vanilla Vision Transformers (ViT) can achieve realistic reconstruction results~\cite{tong2022videomaenju,feichtenhofer2022videomaefb}, which implies that the masked visual information in videos can be complemented from limited visible contents, and the ViT show this capability.
This is possible because that the spatio-temporal redundancy in videos empowers the model to derive the visual semantics of the invisible parts from the visible contexts. 
After pre-training, models are fine-tuned on the downstream action recognition task following a standard scheme, which feeds all the details of videos into ViT. 

\begin{figure}
    \centering
    \includegraphics[width=1.0\linewidth]{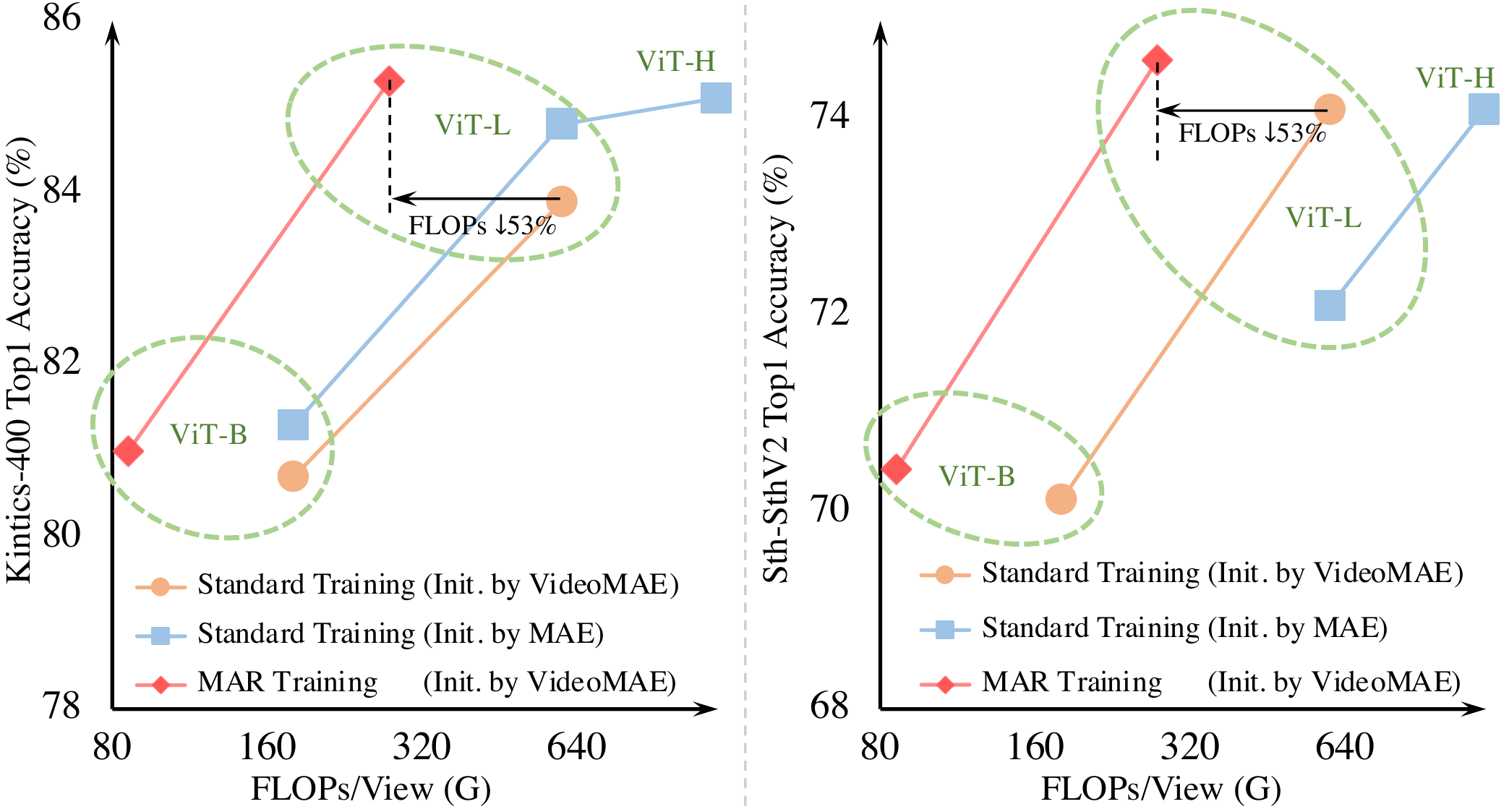}
    \caption{Accuracy vs. per-view GFLOPs on Kinetics-400~\cite{kay2017k400} and Something-Something v2~\cite{goyal2017something}. The model architecture is ViT~\cite{dosovitskiy2020vit}, pre-trained by self-supervised VideoMAE~\cite{tong2022videomaenju} and MAE~\cite{feichtenhofer2022videomaefb}. ``Standard Training'' means using all tokens with a linear classifier for training, which requires more computational resources. 
    The proposed MAR costs only 47\% of the computation to achieve better performance with the same pre-trained encoder.
    }
    \label{fig:first_fig}
    \vspace{-3mm}
\end{figure}

%
%
In this work, we argue that given the strong ability of ViT to reconstruct visual semantics with only limited visual content and the high spatio-temporal redundancy in videos, the standard action recognition scheme that operates full video frames is highly inefficient. To this end, we propose a simple and computationally efficient end-to-end scheme for the action recognition task, termed as Masked Action Recognition (MAR). 
%
The core idea of MAR is to discard a subset of video patches to reduce the encoded tokens of ViT, such that the redundant computation can be avoided to some extent.
%
We investigate this from two perspectives: \textit{(i)} designing an appropriate input masking map with strong spatio-temporal correlations for ViTs; \textit{(ii)} increasing the abstraction level for the features output by ViTs.

For the first perspective, we aim to determine a reasonable mask form for action recognition.
The existing VideoMAE methods~\cite{tong2022videomaenju,feichtenhofer2022videomaefb} adopt a large proportion (\textit{i.e.,} 90\%) of tube masking or random masking, which are mainly designed to avoid the ``information leakage'' caused by spatio-temporal redundancy and correlation in videos to improve the difficulty of the self-supervised reconstruction task.
On the contrary, detailed contexts are crucial for action recognition. 
Therefore, training an accurate action recognizer with masks requires the model to easily complement the disappeared details.
To this end, we present \textit{cell running masking}, which encourages ``spatio-temporal information leakage'' to empower the detail perception of encoders for the lost patches.
Specifically, we first design running masking strategy to drive the masks to move frame by frame so that the patches in the same spatial location can be observed in turn.
To generate more flexible masking maps for training, we divide the running masking into multiple small running cells and place a specified proportion of masks in these cells. 
Executing the running strategy in the small cells can present multiple states to provide various spatio-temporally interleaved masks.
Overall, the cell running masking preserves spatio-temporal correlations~\cite{tong2022videomaenju}, which can be easily exploited by the encoder to perceive missing details for accurate action recognition.
Figure~\ref{fig:vis_reconstruction} shows that the semantically explicit reconstructed videos can be observed with this masking strategy.

Despite this, we also observe that the models with realistic reconstructed videos still fail to achieve on-par accuracy compared to using the full patches.
MAE in the image domain~\cite{he2021mae}
identified a semantic gap between the low-level features required for the reconstruction task and the abstract features required for the recognition task.
Meanwhile, the literature~\cite{wang2022maelite} also found that the models pre-trained by MAE extract features at a lower abstract level.
To solve this, we further propose \textit{bridging classifier} with a similar structure to the reconstruction decoder in MAE.
%
Nevertheless, compared to the reconstruction decoder, the bridging classifier is designed to further bridge the semantic gaps and make the encoded features more specialized for the classification task. In contrast, the reconstruction decoder is used to decode the low-level pixel-wise information.
%

In this way, compared to the standard action recognition scheme, MAR is shown to reduce the computation by up to 53\% while achieving on par or better performances,  as in Figure~\ref{fig:first_fig}.
In particular, when training a large model on Kinetics-400~\cite{kay2017k400}, our ViT-Large costs only 14.5\% of FLOPs and surpasses the performance of the ViT-Huge by 0.2\%.

In a nutshell, our contributions can be summarized as:
\begin{itemize}
    \item
    We explore the masked autoencoders for efficient action recognition, achieving better performance and a 2x wall-clock speedup in training and testing.
    \item 
    We present a novel MAR framework, consisting of cell running masking and bridging classifier to better exploit the spatio-temporal correlations and increase the abstract level of encoded features, respectively.
    \item 
    Extensive experiments on different datasets show that MAR requires only 47\% of the computation to consistently outperform standard action recognition with the same pre-trained models. Especially under comparable computational costs, models trained by MAR significantly outperform the existing state-of-the-art methods.
\end{itemize}

\section{Related Works}
\subsection{Action Recognition}
Action recognition is a fundamental task for video analysis and understanding. 
Recent video networks can be divided into two types, \textit{i.e.,} \textbf{convolution-based} and \textbf{transformer-based}.
The dominant \textbf{convolution-based} methods build upon 3D Convolution Neural Networks (CNNs)~\cite{tran2015c3d,carreira2017i3d,tran2019csn}.
Inspired by the two-stream CNNs, separately modelling spatial appearance and temporal relations by 3D convolutions~\cite{feichtenhofer2019slowfast,wang2018artnet,diba2018stc} have also shown decent accuracy.
However, they suffer from a large computation burden.
To reduce complexity, some approaches attempt to disentangle 3D convolution into spatial and temporal convolution~\cite{tran2018r21d,xie2017s3d,feichtenhofer2020x3d,qiu2017p3d}.
Other works employ 2D convolutions for spatial modelling\cite{wang2016tsn,wang2018tsn} and introduce additional temporal operations, such as calibrating the convolution weights by temporal context~\cite{huang2021tada}, generating adaptive temporal kernels~\cite{liu2021tam}, capturing temporal difference for motion modelling~\cite{wang2021tdn,jiang2019stm,li2020tea}, and shifting part of the channels along the temporal dimension~\cite{lin2019tsm}, \textit{etc}.
Limited by the small receptive field of the convolution operator, these convolution-based methods struggle to model long-range spatial-temporal dependency.
In recent years, the great success of transformers in the image domain~\cite{dosovitskiy2020vit,liu2021swinimage,touvron2021deit} has led to the exploration of \textbf{transformer-based} video networks.
VTN~\cite{neimark2021vtn} adopts ViT~\cite{dosovitskiy2020vit} to extract spatial features, followed by a Longformer~\cite{beltagy2020longformer} to capture temporal relationships.
Both TimeSformer~\cite{bertasius2021timesformer} and ViViT~\cite{arnab2021vivit} factorise different spatial- and temporal- attentions for transformer encoders. 
They suggest that factorised spatial and temporal attention can achieve better performance.
MViT~\cite{fan2021mvit,li2021mvitv2} studies hierarchical transformers with several channel-resolution scale stages, and proposes pooling attention to reduce computation.
In comparison, Video Swin Transformer~\cite{liu2021videoswin} introduces an inductive bias of locality for videos and achieves a better speed-accuracy trade-off.
And Uniformer~\cite{li2022uniformer} captures local spatio-temporal context and global token dependency by convolution in shallow layers and transformer in deep layers, respectively.
All these studies are based on variants of transformer structure.
In this work, the ViTs~\cite{dosovitskiy2020vit} pre-trained by VideoMAE~\cite{tong2022videomaenju} are adopted as our encoder. 
We investigate the input and output of transformer encoders and empirically show that the proposed MAR owns advantages in both efficiency and performance.

\subsection{Spatio-temporal Redundancy}
Reducing spatio-temporal redundancy for efficient video analysis has recently been a popular research topic. 
The mainstream approaches mostly train an additional lightweight network to achieve: \textit{(i)} adaptive frame selection~\cite{wu2019adaframerel3,wu2019rel1,zheng2020dsn,li20212dada3drel5,korbar2019scsamplerrel4}, \textit{i.e.,} dynamically determining the relevant frames for the recognition networks; \textit{(ii)} adaptive frame resolution~\cite{wu2019adaframerel3}, \textit{i.e.,} learning an optimal resolution for each frame online;  \textit{(iii)} early stopping~\cite{fan2018watchingrel7}, \textit{i.e.,} terminating the inference process before observing all frames; \textit{(iv)} adaptive spatio-temporal regions~\cite{wang2021adafocus,wang2021adafocusv2}, \textit{i.e.,}  localizing the most task-relevant spatio-temporal regions; \textit{(v)} adaptive network architectures~\cite{li20212dada3drel5,wu2019liteevalrel2,wu2020dynamicinference}, \textit{i.e.,} adjusting the network architecture to save computation on less informative features.
Another line is to manually define low redundant sampling rules, such as MGSampler~\cite{zhi2021mgsampler}, which selects frames containing rich motion information by the cumulative motion distribution.
Nevertheless, in this work, the ViT trained by MAR adopts only a proportion of video patches for efficient action recognition, which takes advantage of the powerful completion capabilities of transformers.

\begin{figure*}[t]
    \centering
    \includegraphics[width=0.9\linewidth]{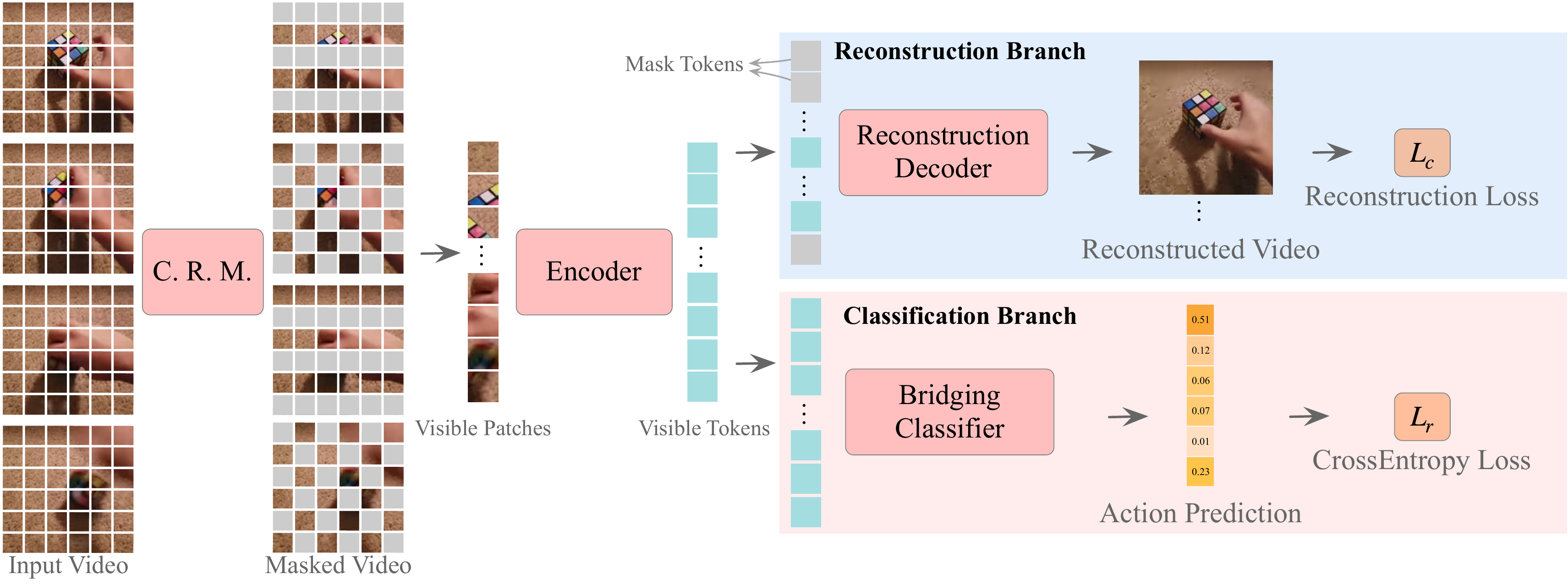}
    \caption{Overview of Masked Action Recognition (MAR). A specified proportion (\textit{e.g.,} 50\% here) of patches is first discarded by cell running masking (C.R.M.), which retains sufficient spatio-temporal correlations. 
    Next, the remaining visible patches are fed into the encoder to extract their spatio-temporal features (\textit{i.e.,} visible tokens in the figure). 
    Finally, the reconstruction decoder receives mask tokens and visible tokens to reconstruct the masked patches. 
    In contrast, the bridging classifier in the classification branch receives only visible tokens from performing the action classification task.
    Note that the reconstruction branch is only performed in training, which is used to preserve the completion capability of the encoders for invisible patches.}
    \label{fig:framework}
    \vspace{-3mm}
\end{figure*}

\subsection{Masked Autoencoders}
Masked autoencoder, as a form of denoising autoencoding~\cite{vincent2008dae}, is a general methodology to learn effective representations by reconstructing the uncorrupted inputs from the corrupted inputs.
In Natural Language Processing (NLP), the masked language modelling task proposed in BERT~\cite{devlin2018bert} is one of the most successful explorations of masked autoencoding.
Various variants~\cite{bao2020unilmv2bert,liu2019roberta,yang2019xlnetbert,dong2019unibert} based on BERT also further improve the performance of language transformer pre-training.
Recently, in the image domain, a series of masked autoencoding methods seek a framework for vision and language unification based on Transformer architectures~\cite{vaswani2017transformer}, and continued progress has been made. 
%
iGPT~\cite{chen2020igpt} first proposes to train a transformer to predict pixels from a sequence of low-resolution pixels for unsupervised representation learning.
Then ViT~\cite{dosovitskiy2020vit} takes image patches as tokens and performs masked patch prediction to mimick the masked language modelling in BERT~\cite{devlin2018bert}.
SimMIM~\cite{xie2022simmim} suggests that a large masked patch size for pixel predictions makes a strong pre-text task.
Image MAE~\cite{he2021mae} investigates an asymmetric encoder-decoder structure, \textit{i.e.,} 
the heavy encoder only operates a small proportion (25\%) of visible patches, while a lightweight decoder is used to reconstruct pixels. 
%
Apart from pixel prediction, another research line also proposes to reconstruct other targets, such as pre-trained dVAE~\cite{van2017dvae1,ramesh2021dvae2} of BEiT~\cite{bao2021beit}, and HoG~\cite{dalal2005hog} of MaskFeat~\cite{wei2022maskedfeat}.
In the video domain, two MAE-based methods~\cite{tong2022videomaenju,feichtenhofer2022videomaefb} find that an extremely high proportion of masks yields decent performance due to the large spatio-temporal redundancy in videos.
Instead of predicting pixels, BEVT~\cite{wang2022bevt} and VIMPAC~\cite{tan2021vimpac} also attempt to learn spatio-temporal representations by predicting features derived from a tokenizer.
All the masked autoencoder based works discussed above focus on learning an effective self-supervised visual representation. 
We are inspired by the powerful completion ability of ViT in reconstruction tasks~\cite{he2021mae,tong2022videomaenju,feichtenhofer2022videomaefb} and propose to adopt this idea to empower supervised action recognition models at both the efficiency and performance levels.

\section{Approach}

\begin{figure*}
    \centering
    \includegraphics[width=1.0\linewidth]{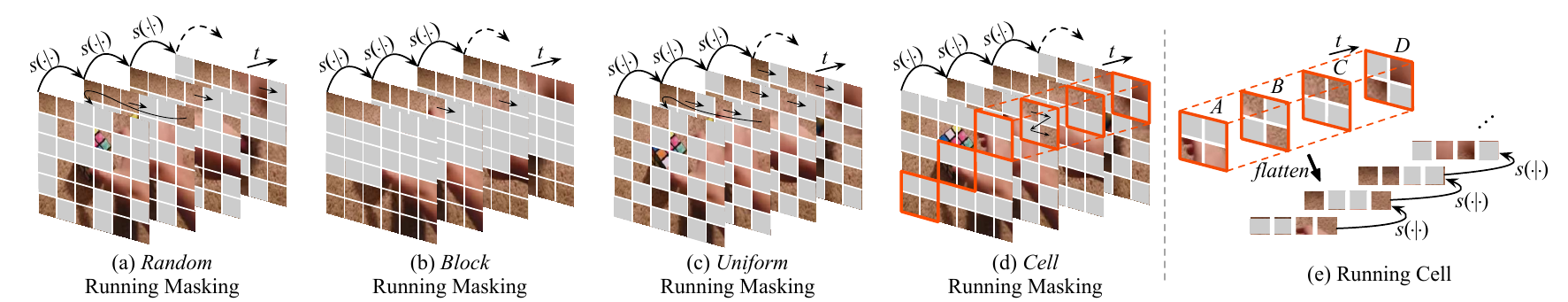}
    \caption{Different schemes for running masking: \textbf{(a)} randomly removing spatial patches in the first frame; \textbf{(b)} removing a large spatial block in the first frame; \textbf{(c)} uniformly distributing the masks in the first frame; \textbf{(d)} performing running masking in cells; \textbf{(e)} the masks run circularly frame by frame in a running cell.}
    \label{fig:mask_comparision}
\end{figure*}
\begin{figure}
    \centering
    \includegraphics[width=0.8\linewidth]{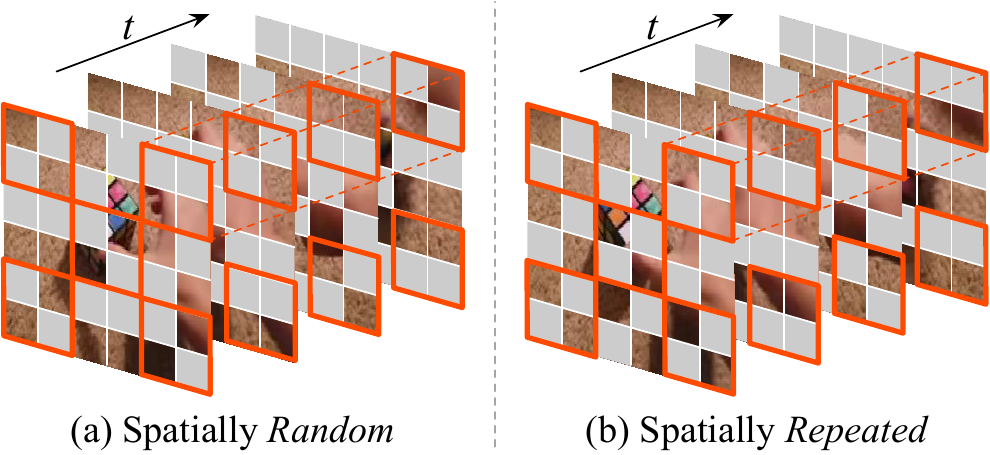}
    \caption{Different combinations of running cells in space: \textbf{(a)} each running cell randomly selects a starting state; \textbf{(b)} all running cells share the same random starting state.}
    \label{fig:running_augmentation}
    \vspace{-3mm}
\end{figure}


%
As shown in Figure 2, 
instead of full video frames, the proposed MAR takes the masked videos as input. 
Specifically, the encoder of MAR only operates over the visible patches, which allows the encoder to process videos with only a fraction of computation and memory cost, \textit{e.g.,} $50\%$ in Figure~\ref{fig:framework}. 
The encoded visible tokens are then fed into two branches: the reconstruction branch for pixel-level reconstruction,
and the classification branch for efficient action recognition.
Note that the auxiliary reconstruction branch is performed only in training and removed in inference.
Besides the general framework, we also present two key designs in MAR:
\textit{(i)} To facilitate the perception of the MAR encoder over invisible patches, we propose a cell running masking strategy (in Sec.~\ref{sec:cell_running_mask}) to generate a masking map that ensures strong spatio-temporal correlations;
\textit{(ii)} To bridge the semantic gaps between the encoded visible tokens and the features that can be better used for action classification task, we introduce a bridging classifier (in Sec.~\ref{sec:briding_classifier}) for the classification branch to increase the abstraction level of tokens.
%
%
%
%
%
%
In this section, we first briefly revisit masked video modelling and then present the idea of cell running masking and bridging classifier.

\subsection{Masked Video Modelling}

The existing masked video modelling approaches with encouraging performances~\cite{tong2022videomaenju,feichtenhofer2022videomaefb} are extended from image MAE.
They first divide an input video into several non-overlapped spatio-temporal patches, and then only a small proportion of patches (\emph{i.e.}, 10\%) with positional embeddings are randomly selected as input for the ViT encoders~\cite{dosovitskiy2020vit}.
A lightweight decoder is then adopted to reconstruct the full video from the encoded latent representations of visible patches.
VideoMAE~\cite{tong2022videomaenju} identifies two crucial characteristics of video data, \textit{i.e.,} \textit{temporal redundancy} and \textit{temporal correlation}. 
The former means that the semantics vary slowly in the temporal dimension, and the spatio-temporal information is highly redundant, indicating that retaining all spatio-temporal patches for training and inference is inefficient and unnecessary.
While the latter emphasizes the strong inherent correlation between adjacent frames, which could cause ``information leakage'' between frames, thus reducing the reconstruction difficulty of VideoMAE.
Hence, a large proportion (\textit{i.e.,} 90\%) of mask ratio and tube masking are proposed for the video reconstruction task.
And the pre-trained models can still achieve satisfactory reconstruction videos with limited visible contents, which implies the powerful spatio-temporal association ability of ViTs.
%
%
%
%
However, in the downstream action recognition task, all patches are still fed into the encoder, especially for a video clip with 1568 tokens, which still causes a heavy computational burden.
In this work, we draw inspiration from the powerful capability of complementing invisible contexts shown by MAE pre-trained ViTs and propose MAR to achieve efficient action recognition.

\subsection{Cell Running Masking}
\label{sec:cell_running_mask}
\noindent
\textbf{Running masking.} 
%
%
When performing the action recognition task with masks, a key issue is that the valuable details in masked patches are removed together, which is bound to damage accuracy.
To this end, instead of avoiding  ``information leakage'' between the adjacent frames like VideoMAE~\cite{tong2022videomaenju}, we encourage ``information leakage'' caused by temporal correlation to lower the difficulty of the reconstruction task
since easier reconstruction means that richer details can be redrawn to improve recognition performance.
%
%
%

Specifically, we first propose running masking strategy, where the masks run frame by frame, and the patches at different spatial locations are discarded in turn.
It can be formulated as:
\begin{equation}
    M_{t} = s(M_{t-1}|M_{t-2},\dots M_{1}),
\end{equation}
where $M_{t}$ is the masking map for frame $t$. Function $s(\cdot|\cdot)$ indices that the masking map of frame $t$ is transformed from frame $t-1$ circularly. 
Figure~\ref{fig:mask_comparision}(e) also shows the masks in a small running cell between adjacent frames.
With this idea, running masking ensures that the same spatial locations in consecutive frames can quickly observe the visible patches from adjacent frames, which exploits spatio-temporal visual redundancy to mitigate information loss.

%
However, it is still a challenge to get an efficient and effective implementation of running masking.
As shown in Figure~\ref{fig:mask_comparision}(a), one possible option is to generate random masks on the first frame and then use the function $s(\cdot|\cdot)$ to transform these masks frame by frame. 
This straightforward idea suffers from the randomness of masks, and patches at mask-dense spatial locations may not be displayed, which still can not effectively exploit video visual correlations.
Similar defects appear in the block-wise running masking illustrated in Figure~\ref{fig:mask_comparision}(b), where the dense masks destroy chunks of spatio-temporal context.
Figure~\ref{fig:mask_comparision}(c) shows a different uniform running masking. 
It arranges uniform grid masks on the first frame to form spatial dislocations that enable the model to infer the spatial semantics for the masked patches with the image spatial redundancy. 
Then these grid masks are shifted in subsequent frames to form temporal intersections, which allows the model to recall the detailed information of the masked parts by temporal correlations in the video.
Hence, the uniform running masking can already efficiently exploit for the spatio-temporal redundancy. 
However, it is still worth noting that adopting the rigid ``uniform masking'' can easily lead to overfitting in training.
Since the diversity of uniform masking maps is limited.

\noindent
\textbf{Running cell.}
For this reason, we propose to decompose the running masking into multiple small repeated units, and one unit is termed a \textit{running cell}. 
The combination of these simple units can also implement complex masks, called \textit{cell running masking}, as shown in Figure~\ref{fig:mask_comparision}(d, e) and Figure~\ref{fig:running_augmentation}(a, b).
Suppose we define the spatial size of a running cell as $r\times q$ (\textit{e.g.,} $r=q=2$ in Figure~\ref{fig:mask_comparision}(e)). 
When the spatial size is small, there are only a few patches in the running cell, and the arrangement of masks is clear. 
For example, in Figure~\ref{fig:mask_comparision}(e), only two masks need to be placed, when the mask ratio is set to 50\%, 
We simply put the two masks in the first two patches and define this as state $A$.
Driven by the function $s(\cdot|\cdot)$, there are four different states, \textit{i.e.,} $A\rightarrow B\rightarrow C\rightarrow D$, in this cell.
And the cell starts with state $A$ and performs state switching with a period of 4 in temporal dimension to realize a spatio-temporal uniform masking map.
This design ensures that each patch has an equal chance of being visible in four temporal patches, thus providing enough temporal correspondences for other masked patches with the same spatial locations.

We also observe that a running cell with a spatial size of $2\times 2$ can only provide three different mask ratios, \textit{i.e.,} 25\%, 50\%, and 75\%, which already meets most of the needs. 
Using running cells with a large spatial size can produce more mask ratios, but the diversity of mask combinations is weakened since there are fewer cells.
For example, when the size of the running cell is equal to the spatial size of the ViT encoded tokens, \textit{i.e.,} $14\times 14$, it degenerates to uniform running masking.

\noindent
\textbf{Augmentations of cell running masking.}
Figure~\ref{fig:mask_comparision}(e) shows that there are four different states in turn, \textit{i.e.,} $A\rightarrow B\rightarrow C\rightarrow D$, in the illustrated running cell.
Clearly, the fixed sequence and positions may cause overfitting in training, similar to the uniform running masking in Figure~\ref{fig:mask_comparision}(c). 
However, with the design of the running cell, we can put multiple different cells in the space, and these cells are free to choose the starting states. 
As shown in Figure~\ref{fig:running_augmentation}(a), each running cell selects the starting state randomly, and their combination is close to the random masking, but the regulation of running masking is followed within each small cell.
Figure~\ref{fig:mask_comparision}(d) and Figure~\ref{fig:running_augmentation}(b) repeat running cells spatially, but their starting states are selected randomly as $A$ and $B$, respectively, resulting in visually different masking maps.
Both the spatially random and spatially repeated masking in Figure~\ref{fig:running_augmentation} can be used as a data augmentation in training. 
In inference, we simply use the spatially repeated running masking for evaluation. 
The experiments in Table~\ref{tab:mask_aug_starting_state_cross_val} (b) demonstrate that the model is robust to the different starting states of running cells in inference.

\subsection{Bridging Classifier}
\label{sec:briding_classifier}
The design of cell running masking preserves the spatio-temporal correlations between visible patches and invisible patches, effectively reducing the difficulty of reconstruction and further improving the reconstruction quality.
For example, when the mask ratio is set to 50\%, the reconstruction branch can already achieve sufficiently satisfactory reconstructed videos, as shown in Figure~\ref{fig:vis_reconstruction}. 
However, when using a fully connected layer as a classifier in Table~\ref{tab:mask_ratio_decoder_head}, the recognition accuracy with 50\% of the masked patches still struggles to reach the performance without masking. 
Our bridging classifier sets out to close this performance gap.

MAE~\cite{he2021mae} mentioned that the pixel-level reconstruction and the recognition tasks require latent representations at different abstract levels.
Specifically, representations with higher-level semantics can lead to better recognition accuracy but are not specialized for reconstruction.
In this work, the satisfactory reconstructions prove that the low-level semantic information in features is sufficient. Hence, we speculate that the weak classification accuracy is actually because the linear classifier cannot fully exploit low-level semantic information.

To this end, we propose \textit{bridging classifier}, consisting of a series of transformer blocks like the reconstruction decoder,
to bridge the semantic gaps between the encoded features and the classification features:
\begin{equation}
    \mathbf{p}=\phi(h(\mathbf{F}_v)),
\end{equation}
where $\mathbf{F}_v\in R^{N_v\times D}$, is the encoded visible tokens output by the encoder. $\phi(\cdot)$ and $h(\cdot)$ indicate average pooling operation and proposed bridging classifier. $N_v$ is the number of visible tokens, and $D$ is the channel. $\mathbf{p}\in R^{C}$ is the model prediction for $C$ classes.
The reconstruction decoder needs to infer the encoded $F_v$ to pixels, while the bridging classifier only extracts the more concise classification features in $F_v$. 
Therefore, the bridging classifier is designed to be more lightweight.
More importantly, the bridging classifier only processes visible tokens, which only costs 8\% FLOPs \textit{vs.} linear classifier with 50\% of visible tokens.

\subsection{Loss Function}
MAR contains the reconstruction and classification branches, and their loss functions are denoted as  $\emph{L}_r$ and $\emph{L}_c$, respectively. 
The reconstruction loss $\emph{L}_r$ is pixel-wise mean-squared loss, following~\cite{tong2022videomaenju}. And $\emph{L}_c$ is a widely used cross entropy loss in classification tasks.
The training objective function can be written as follows:
\begin{equation}
    \emph{L}_r = \frac{1}{\Omega(\mathbf{x}_M)}||\mathbf{x}_M-\mathbf{y}_M||_2 \\
\end{equation}
\begin{equation}
    \emph{L}_c = -\sum_{i=1}^C\mathbf{z}_i\text{log}(\mathbf{p}_{i}) \\
\end{equation}
\begin{equation}
     \emph{L} = \lambda \emph{L}_r + \emph{L}_c,
    \label{eq:loss}
\end{equation}
where $\mathbf{x},\mathbf{y}$ are the input RGB pixel values and the predicted pixel values, respectively; $M$ denotes the masked pixels; $\Omega(\cdot)$ calculates the number of masked pixels; $||\cdot||_2$ refers to $\ell_2$ loss; $\mathbf{z}$ is a one-hot label vector for classification; $\lambda$ is the balance parameter for the reconstruction branch.

\begin{table}[!t]
    \centering
    \small
    \caption{\textbf{MAR training settings}. \label{tab:ft_settings}}
    \begin{tabular}{c|cc}
        Config & Sth-Sth v2 & Kinetics-400 \\
        \midrule[1.15pt]
        Optimizer & \multicolumn{2}{c}{AdamW~\cite{loshchilov2017adamw}} \\
        Momentum & \multicolumn{2}{c}{$\beta_1=0.9,\beta_2=0.999$} \\
        Weight Decay & \multicolumn{2}{c}{0.5} \\
        Base LR$^\dag$  & \multicolumn{2}{c}{1e-3} \\
        Batch Size & \multicolumn{2}{c}{512(B),128(L)}\\
        LR Schedule  & \multicolumn{2}{c}{cosine decay~\cite{loshchilov2016sgdr}} \\
        Layer-wise Decay  & \multicolumn{2}{c}{0.75} \\
        Filp Augmentation & \multicolumn{2}{c}{yes} \\
        RandAugment~\cite{cubuk2020randaugment} & \multicolumn{2}{c}{(9, 0.5)} \\
        Mixup~\cite{zhang2017mixup} & \multicolumn{2}{c}{0.8} \\
        Cutmix~\cite{yun2019cutmix} & \multicolumn{2}{c}{1.0} \\
        Label Smoothing~\cite{szegedy2016labelsmooth} & \multicolumn{2}{c}{0.1} \\
        Droppath~\cite{huang2016droppath} & \multicolumn{2}{c}{0.1(B), 0.2(L)} \\
        Dropout~\cite{srivastava2014dropout} & \multicolumn{2}{c}{0.1} \\
        Warmup Epochs~\cite{goyal2017warmup}  & \multicolumn{2}{c}{5} \\
        Training Epochs  & 40 & 100(B),60(L) \\
        Repeated Sampling~\cite{hoffer2020batchaug} & 1 & 2 \\
    \end{tabular}
    \begin{tablenotes}
    \footnotesize
    \item $^\dag$ is linear learning rate scaling rule~\cite{goyal2017warmup}:  $\text{ActualLR}=\text{BaseLR}\times\text{BatchSize}/256$.
    \end{tablenotes}
\vspace{-3mm}
\end{table}

\section{Experiments}
\subsection{Implementation}
\noindent
\textbf{Datasets.}
We evaluate MAR on four widely-used action recognition datasets, \textit{i.e.,} Kinetics-400~\cite{kay2017k400}, Something-Something v2~\cite{goyal2017something}, HMDB51~\cite{kuehne2011hmdb}, and UCF101~\cite{soomro2012ucf101}. Kinetics-400 is a large-scale benchmark with around 240k training videos and 20k validation videos from 400 different action categories. 
Something-Something v2 is a temporal-related video dataset with 174 action classes, which contains 169k videos for training and 20k videos for validation. HMDB51 and UCF101 are two small datasets for action recognition, which only have 3.5k/1.5k train/val videos and 9.5k/3.5k train/val videos, respectively.

\noindent
\textbf{Architecture.}
Following VideoMAE~\cite{tong2022videomaenju} and MAE~\cite{feichtenhofer2022videomaefb}, we use ViT~\cite{dosovitskiy2020vit} with the joint spatial-temporal attention.
The attention mechanism with quadratic complexity can lead to computational bottlenecks, while MAR can effectively alleviate this problem.
%
%
We set the same spatio-temporal patch size (\textit{i.e.,} $2\times 16\times 16$) as VideoMAE for both ViT-Base and ViT-Large to utilize its pre-trained models conveniently.
The spatio-temporal resolution of input videos is $16\times 224 \times 224$ for both training and inference, and the embedding tokens output by encoders are $8\times 14\times 14=1568$. 
When the mask ratio is set to 50\%, the encoders only operate 784 tokens.

\noindent
\textbf{Data pre-processing and training settings.}
Our MAR training configurations follow the training settings in VideoMAE~\cite{tong2022videomaenju}. 
We use the same training augmentations as VideoMAE, as shown in Table~\ref{tab:ft_settings}.

\begin{table}[!t]
\caption{\textbf{Mask sampling.} ``$\rho$'' is the mask ratio. ``D.S. 2x'' means that the spatial resolution is downsampled from $224^2$ to $112^2$. 
\label{tab:mask_sampling}}
\centering
\small
\tablestyle{4pt}{1.0}
\begin{tabular}{cccccc}
$\rho$ & \multicolumn{2}{c}{Masking Strategy} & Top-1 & Top-5 & GFLOPs\\
\midrule[1.15pt]
\multirow{2}{*}{25\%} & Random &  Standard &  71.12 & \textbf{93.26} &  138.04\\ 
~& Cell  &  \textit{Running} & \textbf{71.30}  &93.15& 138.04\\ 
\midrule
\multirow{8}{*}{50\%} &Block &  Standard & 64.73 &89.37&86.35\\ 
~ &Block &  \textit{Running} &65.00 &89.74&86.35\\ 
\cmidrule(r){2-6}
~ &Random & Standard &70.35 &92.63&86.35\\ 
~ &Random & \textit{Running}&70.46 &92.74&86.35\\ 
\cmidrule(r){2-6}
~ &Frame &  Standard &  66.25 & 89.85& 86.35\\ 
~ &Tube  &  Standard &  70.25 & 92.73 & 86.35\\ 
~ &Uniform& \textit{Running}&  70.55 & \textbf{92.84} &  86.35\\ 
~ &\cellcolor{grey}Cell& \cellcolor{grey}\textit{Running} & \cellcolor{grey}\textbf{70.97} & \cellcolor{grey}92.75 & \cellcolor{grey}86.35\\ 
\midrule
0\% & D.S. 2x & - &  67.23 & 90.85 & 39.56\\ 
\midrule
\multirow{2}{*}{75\%} & Random & Standard &  68.79 & 91.55 & 40.95\\ 
~ &Cell  &  \textit{Running} &  \textbf{69.46} & \textbf{91.88} & 40.95\\ 
\end{tabular}
\end{table}

\subsection{Ablation Studies}
In this section, we present ablation studies for the in-depth analysis of our proposed complements (\textit{i.e.,} cell running masking and bridging classifier) in MAR. 
Something-Something v2~\cite{goyal2017something} is employed as our evaluation benchmark. 
The default encoder is ViT-Base~\cite{dosovitskiy2020vit} with 16 frames, and other default settings are marked in \colorbox{baselinecolor}{gray} in tables. The notation $\rho$ in tables means the mask ratio. 
If not specific, the mask ratio is set to 50\% by default.

\noindent
\textbf{Different mask sampling strategies.} We replace the cell running masking in MAR with other mask sampling methodologies under different mask ratios $\rho$ Table~\ref{tab:mask_sampling}. Several observations can be summarized  from the table: 
\textit{(i)} Compared with ``random standard masking'', our ``cell running masking'' can provide stable accuracy improvements. 
Especially, more gains can be yielded by cell running masking with larger the mask ratios, \textit{e.g.,} the gains are 0.18\%, 0.62\% and 0.67\% for mask ratios of 25\%, 50\% and 75\%, respectively. 
This can be explained by the less destroyed contexts with the small mask ratio,
and cell running masking can better show its advantages with a large proportion of masks;
\textit{(ii)} ``block standard masking'' and ``frame standard masking'' remove chunks of spatio-temporal information and destroy the natural spatio-temporal correlations in videos, significantly impairing performance.
%
%
Nevertheless, improvements are still observed from ``block standard masking'' to ``block running masking'', which demonstrates the effectiveness of our running masking strategy;
\textit{(iii)} Downsampling is also a straightforward way to reduce the computational costs.
We downsample the width and height of the input videos to half of the standard training settings, \textit{i.e.,} $16\times 112^2$, and the computation is comparable to that when we set the mask ratio to 75\%.
It can be observed that downsampling notably degenerates the performance by 2.23\% from cell running masking, which indicates that the low resolution drops more detailed information than masks;
\textit{(iv)} Compared with ``random standard masking'', ``uniform running masking'' without running cells has limited improvement, while our `cell running masking' performs much stronger. 
The results validate that our cell running masking can benefit from the diversity brought by various cell states in training.

\begin{table*}[t]
\caption{\textbf{(a) Masking augmentations.} Spatial and temporal augmentations of cell running masking in training. ``Random'' and ``Repeated'' are two cell combinations shown in Figure~\ref{fig:running_augmentation}. ``Fixed'' means the fixed running state order. ``Shuffled'' indicates the shuffled order. \textbf{(b) Starting states in testing.} Running cells with varying starting states for model inference. \textbf{(c) Cross validations of different mask ratios.} Here only Top-1 accuracy is reported. \label{tab:mask_aug_starting_state_cross_val}}
\centering
\subfloat{
\centering
\begin{minipage}{0.3\linewidth}{\begin{center}
\tablestyle{4pt}{1.05}
\begin{tabular}{ccccc}
Spatially & Temporally & Top-1 & Top-5 \\
\midrule[1.15pt]
Random & Fixed & 70.61 & 92.91 \\ 
Random & Shuffled & 70.58 & 92.95 \\  
Repeated & Fixed & 70.72 & \textbf{92.96}\\ 
\cellcolor{grey}Repeated & \cellcolor{grey}Shuffled &\cellcolor{grey}\textbf{70.97}  & \cellcolor{grey}92.75\\
\multicolumn{4}{c}{(a)}\\
\end{tabular}
\end{center}}\end{minipage}
}
\hspace{0em}
%
\subfloat{
\begin{minipage}{0.22\linewidth}{\begin{center}
\tablestyle{4pt}{1.05}
\centering
\begin{tabular}{cccc}
Start. State & Top-1 & Top-5\\
\midrule[1.15pt]
\cellcolor{grey}$\mathbf{A}BCD$ & \cellcolor{grey}70.97 & \cellcolor{grey}92.75 \\
$\mathbf{B}CDA$ & \textbf{70.98} & 92.70 \\
$\mathbf{C}DAB$ & 70.97 & \textbf{92.79} \\
$\mathbf{D}ABC$ & 70.95 & 92.77\\
\multicolumn{3}{c}{(b)}\\
\end{tabular}
\end{center}}
\end{minipage}
}
\hspace{0em}
%
\subfloat{
\begin{minipage}{0.38\linewidth}{\begin{center}
\tablestyle{4pt}{1.05}
\centering
\begin{tabular}{ccccccc}
\multicolumn{2}{c}{Training$\rightarrow$}& 0\% & 25\% & 50\% & 75\% & GFLOPs\\
\cmidrule[1.15pt](r){1-7}
\multirow{4}{*}{\rotatebox{-90}{Testing}} & 0\% & 71.02 &\textbf{71.36} &71.32&70.17& 196.03 \\
&25\% & 70.35 & 71.30 & 71.22  & 70.09& 138.04\\
&50\% & 69.37 & 70.94 &\cellcolor{grey}70.97&70.09& 86.35\\
&75\% & 66.31 & 68.29 &69.49&69.46& 40.95\\
\multicolumn{7}{c}{(c)}\\
\end{tabular}
\end{center}}
\end{minipage}
}
\vspace{-7mm}
\end{table*}

\begin{table}[!t]
\caption{\textbf{The spatial size of running cell.} ``$r\times q$'' is the spatial size described in Sec.~\ref{sec:cell_running_mask}. \label{tab:cell_size}}
\centering
\small
\begin{tabular}{c|ccccc}
$r\times q$  & $14\times 14$ & $7\times 7$&  $7\times 2$ & $2\times 7$ & \cellcolor{grey}$2\times 2$\\
\midrule[1.15pt]
Top-1  & 70.55 & 70.54 & 70.65 & 70.84 &\cellcolor{grey} \textbf{70.97}\\
\end{tabular}
\vspace{-3mm}
\end{table}

\begin{figure}[t]
    \centering
    \includegraphics[width=1.0\linewidth]{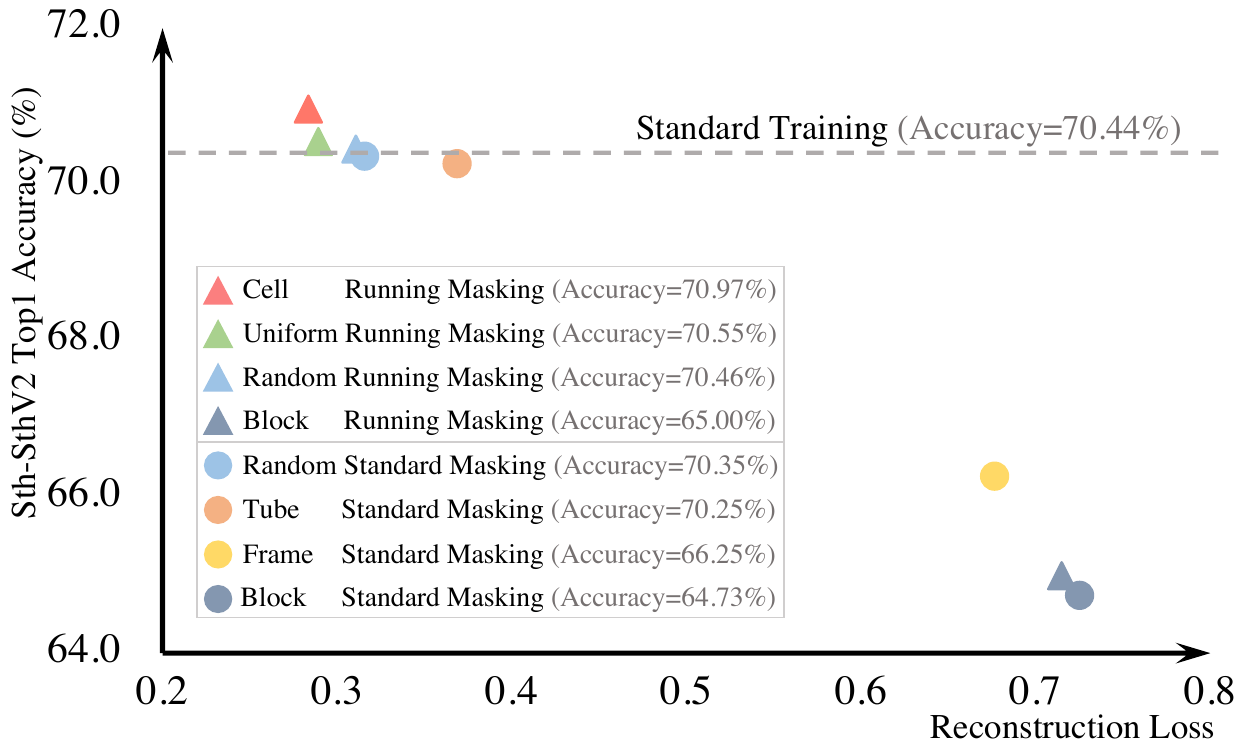}
    \caption{Accuracy vs. reconstruction loss on Something-Something v2. The mask ratio is 50\%, except that the standard training scheme uses all patches as input (\textit{i.e.,} no mask). There is a clear negative correlation between reconstruction loss and accuracy.}
    \label{fig:mae_loss_acc}
    \vspace{-3mm}
\end{figure}

\noindent
\textbf{The spatial size of running cell.} Next, we show that our defined running cell plays a critical role in running masking. 
As in Table~\ref{tab:cell_size}, the running masking with a large cell size, \textit{e.g.,} $14\times14$ or $7\times7$, shows weaker performance than a small cell size, \textit{e.g.,} $2\times2$. 
This is caused by the unitary form of masks.
Specifically, to prevent the masks in the large cell from degenerating into a block-wise mask, the first frame should be uniformly initialized. 
For example, the cell size of uniform running masking in Figure~\ref{fig:mask_comparision}(c) is $14\times14$.
Hence, as discussed in Sec.~\ref{sec:cell_running_mask}, large cell sizes with uniform masks suffer from limited diversities and lead to overfitting in training. 
Our small cell size ($2\times2$) is more flexible and achieves the highest performance.

\noindent
\textbf{Reconstruction loss and accuracy.} 
Here, we show the correlation between reconstruction loss and accuracy in Figure~\ref{fig:mae_loss_acc}.
We can draw two intriguing findings.
First, the accuracy is inversely related to reconstruction loss.
For example, cell running masking can achieve the highest accuracy with the lowest reconstruction loss, while other mask sampling strategies with larger reconstruction loss maybe yield weaker performances.
This can be interpreted as a small reconstruction loss that can recover rich details for invisible patches and thus boost the classification accuracy.
Second, the block masking and the random masking with our running strategy can reduce the reconstruction error and improve the accuracy by $\sim $ 0.3\%, demonstrating that the encoders can exploit more spatio-temporal correlations from proposed running strategy.
%


\noindent
\textbf{Training augmentations of cell running masking.} One of the advantages of the running cell with a small spatial size is its flexibility, \textit{i.e.,} a simple combination of multiple cells can achieve a complex mask. 
We further ablate different spatial and temporal augmentations in training for cell running masking in Table~\ref{tab:mask_aug_starting_state_cross_val}(a).
``Spatially Random'' means that each cell chooses starting state randomly, as shown in Figure~\ref{fig:running_augmentation}(a).
We observe that the random combinations of the cells converge to random masking, and therefore show poor performance. 
``Spatially Repeated'' indicates that the cells share the same randomly selected starting state, as shown in Figure~\ref{fig:mask_comparision}(c) and  Figure~\ref{fig:running_augmentation}(b), demonstrating a better performance. 
The temporal shuffling slightly increases the randomness and thus the best performance.
This implies that the training masks with reasonable randomness can introduce about 0.4\% improvements, and the simple control of running cells in the spatial and temporal space achieves this regulation.

\begin{table}[!t]
\caption{\textbf{Bridging classifier with different mask ratios.} ``$\rho$'' is the mask ratio and $\Delta$ is an absolute improvement of Top1 accuracy. ``GPU Mem.'' indicates the usage of GPU memory in training when the batch size is set to 1. \label{tab:mask_ratio_decoder_head}}
\centering
\small
\begin{tabular}{cccccc}
$\rho$& Classifier  & Top-1 & Top-5 & $\Delta$ & GPU Mem.\\
\midrule[1.15pt]
0\% & Linear   &  70.44 & 92.67  & - & 14.15 G\\  
0\% & Bridging  & \textbf{71.02} & \textbf{93.18} & +0.58 & 15.08 G \\ 
\midrule
25\% & Linear  & 70.25 & 92.50 &- & 10.54 G\\ 
25\% & Bridging  & \textbf{71.30} & \textbf{93.15} & +1.05 & 11.21 G \\  
\midrule
50\% & Linear  & 69.94 & 92.48 & - & 7.31 G\\ 
\cellcolor{grey}50\% & \cellcolor{grey}Bridging & \cellcolor{grey}\textbf{70.97} & \cellcolor{grey}\textbf{92.75} & \cellcolor{grey} +1.03 & \cellcolor{grey}7.63 G\\ 
\midrule
75\% & Linear  & 68.43 & 91.47 & - & 5.01 G\\
75\% & Bridging  & \textbf{69.46} & \textbf{91.88} & +1.03 & 5.21 G \\ 
\end{tabular}
\vspace{-3mm}
\end{table}

\noindent
\textbf{Different starting states in testing.} There are four different running states in our proposed running cell with a spatial size of $2\times 2$. 
Different starting states will show different masking maps when evaluating the trained models.
It is also worth evaluating the impact of different starting states on performance.
As shown in Table~\ref{tab:mask_aug_starting_state_cross_val}(b), the different starting states present comparable performance, and their differences are negligible. 
This proves that the models trained by MAR are robust to different states of cell running masking.

\noindent
\textbf{Cross validations of different mask ratios.} Table~\ref{tab:mask_aug_starting_state_cross_val}(c) presents the accuracies with different mask ratios for training and testing. 
We can draw the following observations: \textit{(i)} When testing the models with a large mask ratio, \textit{e.g.,} 75\%, the models also should be trained with the akin mask ratios. Otherwise, the model cannot learn to complement the lost contexts;
\textit{(ii)} appropriately increasing the mask ratio in training can also improve accuracy. 
For example, training with 25\% masks and testing with no mask has an advantage over the training with no mask by 0.34\%, \textit{i.e.,} 71.36\% \textit{vs.} 71.02\%;
\textit{(iii)} large mask ratios can remarkably save computational costs. Our default setting with only 50\% masks can save more than 50\% of computation while preserving comparable performance to that when training and testing with no mask. Thus our proposed MAR saves both training and testing overhead.

\noindent
\textbf{The effect of bridging classifier.} The comparisons of the proposed bridging classifier and linear classifier under different mask ratios are ablated in Table~\ref{tab:mask_ratio_decoder_head}. 
It can be observed that larger mask ratios can lead to more performance degradation. 
While simply replacing the linear classifier with a bridging classifier brings notable improvements. 
Especially for large mask ratios, the bridging classifier has an advantage over the linear baseline of around 1.0\%.
This suggests that the mask exacerbates the nonlinearity of the features and confirms our above statement that the abstraction level of encoded features still needs to be further bridged for classification.

\begin{table}[t]
\small
\caption{\textbf{The design of bridging classifier.} ``MLP'' means Multi-Layer Perceptron. \label{tab:decoder_design}}
\centering
\begin{tabular}{cccccc}
Width & Classifier & Depth & Top-1 & Top-5 & GFLOPs\\
\midrule[1.15pt]
-& Linear & - & 69.94 & 92.48 & 79.84\\ 
512& MLP    & 2 & 69.90 & 92.30 & 80.15\\ 
\midrule
\multirow{4}{*}{256}& \multirow{4}{*}{Bridging} & 1 & 70.50 &  92.62& 80.93 \\
&  & 2 & 70.61 & 92.76 & 81.86 \\
&  & 4 & 70.74 & 92.75 & 83.73 \\ 
&  & 8 & 70.78 & 92.86 & 87.46 \\ 
\midrule
\multirow{3}{*}{384}& \multirow{3}{*}{Bridging} & 1 & 70.67 & 92.73& 81.93 \\ 
&  & 2 & 70.89 & 92.89 & 83.80 \\
&  & 4 & 70.76 & 92.73 & 87.52\\ 
\midrule
\multirow{3}{*}{512}& \multirow{3}{*}{Bridging} & 1 & 70.67 & 92.78& 83.25 \\ 
&  & \cellcolor{grey}2 & \cellcolor{grey}\textbf{70.97} & \cellcolor{grey}92.75 & \cellcolor{grey}86.35 \\ 
&  & 4 & 70.78 & \textbf{92.91} & 92.55\\ 
\midrule
\multirow{3}{*}{640}& \multirow{3}{*}{Bridging} & 1 & 70.72 & 92.72 & 84.87 \\ 
&  & 2 & 70.25 & 92.65 & 89.52 \\ 
&  & 4 & 69.05 & 91.94 & 98.81\\ 
\end{tabular}
\vspace{-3mm}
\end{table}

\noindent
\textbf{The width and depth design of bridging classifier.} The lightweight bridging classifier consists of multiple vanilla transformer~\cite{vaswani2017transformer} blocks.
Table~\ref{tab:decoder_design} shows the impact of different decoder designs on the performance.
First, Multi-Layer Perceptron (MLP) is widely known to be a simple design with nonlinear modelling capability. 
However, its performance is slightly weaker than that of linear classifiers.
This indicates that the simple structure of MLP can not squeeze out abstract information from the encoded features.
In contrast, a notable improvement of 0.56\% is observed with only an extremely lightweight decoder, \textit{i.e.,} width=256 and depth=1, which shows the superior nonlinear expressiveness of the transformer block.
Enlarging the width and depth can further boost the performance, with width=512 and depth=2 having the strongest performance.
Wider or deeper decoders not only lead to overfitting but also tend to introduce a more unnecessary computational burden.
Finally, since the decoder only operates visible tokens, the setting with the highest performance adds only 6.51 GFLOPs (around 8\%) compared to the linear classifier.

\begin{table}[!t]
\caption{\textbf{The input of bridging classifier.} ``Pos.Emb.'' is positional embeddings. \label{tab:cls_decoder_input}}
\centering
\small
\begin{tabular}{ccccc}
Pos.Emb. & Masked Tokens & Top-1 & Top-5 & GFLOPs\\
\midrule[1.15pt]
\cmark &  \xmark & 70.66 & 92.78 &86.35 \\ 
 \cmark &  \cmark & 70.44 & \textbf{92.79}&95.06\\ 
 \cellcolor{grey}\xmark & \cellcolor{grey}\xmark &\cellcolor{grey}\textbf{70.97} &\cellcolor{grey}92.75 &\cellcolor{grey}86.35\\
\end{tabular}
\end{table}


\noindent
\textbf{The input of bridging classifier.} MAE-based approaches~\cite{he2021mae,tong2022videomaenju,feichtenhofer2022videomaefb} skip the masked tokens in the encoder and apply them with positional embeddings in the lightweight reconstruction decoder. 
We also evaluate their designs in our proposed bridging classifier in Table~\ref{tab:cls_decoder_input}.
It can be summarized that both the additional positional embeddings and the masked tokens damage the accuracy.
This is probably because both factors are low-level priors and thus more specialized for the reconstruction task. 
In contrast, the high-level semantics required by the classification task is not strongly dependent on these two factors.


\begin{table}[!t]
\caption{\textbf{Reconstruction target and the initialization of reconstruction decoder (Re.Decoder).} \label{tab:mae_target_init}}
\centering
\small
\tablestyle{4pt}{1.0}
\begin{tabular}{ccccc}
Target & Initialization of Re.Decoder & Top-1 & Top-5\\
\midrule[1.15pt]
Pixels w/o norm &  Random & 70.69 & 92.61\\
Pixels w/o norm &  VideoMAE Pre-trained~\cite{tong2022videomaenju} & 70.71 & \textbf{93.04} \\ 
Pixels w/ norm &  Random & 70.82 & 92.83\\ 
\cellcolor{grey}Pixels w/ norm &  \cellcolor{grey}VideoMAE Pre-trained~\cite{tong2022videomaenju} &\cellcolor{grey}\textbf{70.97} &\cellcolor{grey}92.75\\
\end{tabular}
\vspace{-3mm}
\end{table}

\noindent
\textbf{Reconstruction target.} MAE in the image domain~\cite{he2021mae} and video domain~\cite{feichtenhofer2022videomaefb} both demonstrate that reconstructing per-patch normalized pixels works well for self-supervised pre-training. 
As shown in Table~\ref{tab:mae_target_init}, we are interested in whether this finding still holds in MAR.
Compared with reconstructing the original video pixels, using normalized pixels~\cite{he2021mae,feichtenhofer2022videomaefb} always performs better, which is in line with the pre-training settings.

\noindent
\textbf{The initialization of reconstruction decoder.}
We utilize VideoMAE~\cite{tong2022videomaenju} pre-trained parameters to initialize the ViT encoder and the reconstruction decoder by default. 
The effect of the initialization state of the reconstruction decoder is explored in Table~\ref{tab:mae_target_init}. 
It can be observed that the randomly initialized decoder performs decently, while the pre-trained decoder still improves by 0.15\%.
We speculate that the pre-trained decoder has already converged, which can regularize the encoder directly.

\begin{table}[t]
\caption{\textbf{(a) Reconstruction loss weight.} ``$\lambda$'' is the balance parameter in Equation~\ref{eq:loss}. \textit{Note that the reconstruction branch does not work when $\lambda=0$.} \textbf{(b) Pre-training Datasets.} \label{tab:re_w_pt_ds}}
\vspace{-5mm}
\centering
\subfloat{
\centering
\begin{minipage}{0.4\linewidth}{\begin{center}
\tablestyle{4pt}{1.05}
\begin{tabular}{cccc}
$\lambda$ & Top-1 & Top-5\\
\midrule[1.15pt]
0.0 & 70.72 & 92.79 \\
\cellcolor{grey}0.1 & \cellcolor{grey}\textbf{70.97} & \cellcolor{grey}92.75 \\
0.2 & 70.82 & 92.80 \\
0.4 & 70.82 & \textbf{92.93} \\
1.0 & 70.78 & 92.72 \\
\multicolumn{3}{c}{(a)}\\
\end{tabular}
\end{center}}\end{minipage}
}
\hspace{0em}
%
\subfloat{
\begin{minipage}{0.5\linewidth}{\begin{center}
\tablestyle{4pt}{1.05}
\centering
\begin{tabular}{ccc}
Dataset & Top-1 & Top-5\\
\midrule[1.15pt]
None  & 44.37 & 73.18 \\
IN-1K~\cite{deng2009imagenet} & 64.29 & 88.34 \\
K400~\cite{kay2017k400} & 70.49 & 92.66 \\
\cellcolor{grey}SSv2~\cite{goyal2017something} & \cellcolor{grey}\textbf{70.97} & \cellcolor{grey}\textbf{92.75} \\
\multicolumn{3}{c}{~}\\
\multicolumn{3}{c}{(b)}\\
\end{tabular}
\end{center}}
\end{minipage}
}
\vspace{-5mm}
\end{table}

\noindent
\textbf{The effect of reconstruction loss weight, \textit{i.e.,} $\lambda$.} 
%
The reconstruction branch is only involved in the computation in training, and its main purpose is to further enhance the encoder's ability to perceive the missing context by reconstructing the invisible patches. 
In Table~\ref{tab:re_w_pt_ds}(a), we analyze the parameter sensitivity of $\lambda$.
When $\lambda$ is set to 0.0, no reconstruction branch shows the weakest performance.
We observe that a slight increase in $\lambda$, \textit{i.e.,} 0.1, brings improvement. 
Larger $\lambda$, \textit{i.e.,} 1.0 with the stronger reconstruction constraint, may lead the encoder to preserve more low-level cues in features, also resulting in a 0.19\% performance degradation.


\begin{figure*}[t]
    \centering
    \includegraphics[width=1\linewidth]{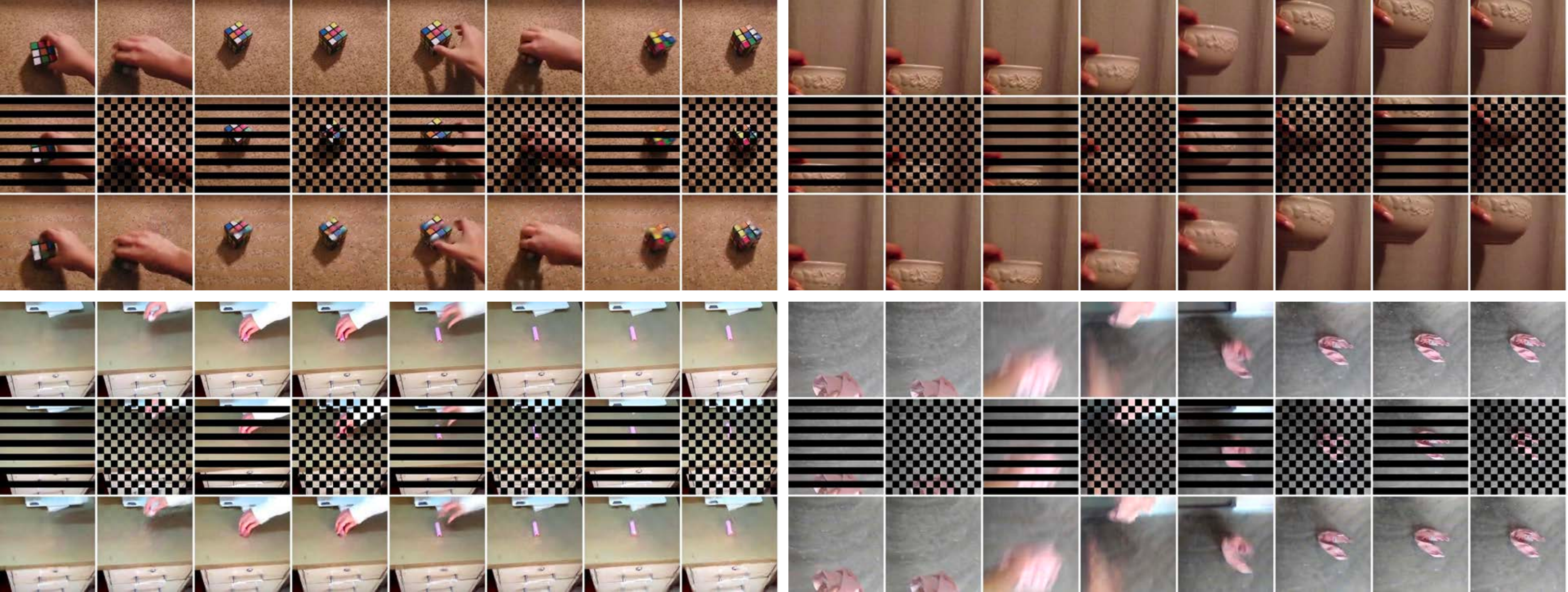}
    \caption{Reconstructed videos of our reconstruction branch on Something-Something v2~\cite{goyal2017something} validation videos. The mask ratio is 50\%. For each video, we show the original video (top), masked video (middle) and reconstructed video (bottom). The model reconstructs the original pixels.}
    \label{fig:vis_reconstruction}
    \vspace{-3mm}
\end{figure*}

\begin{table}[!t]
\caption{\textbf{Data augmentations.} ``RandAug.'' and ``RandEra.'' are RandAugment and random erasing, respectively. \label{data_aug}}
\centering
\small
\begin{tabular}{cccccc}
\makecell{RandAug.\\ \cite{cubuk2020randaugment}} & \makecell{RandEra.\\ \cite{zhong2020randomerasing}} & \makecell{MixUp\\ \cite{zhang2017mixup}} & \makecell{CutMix\\ \cite{yun2019cutmix}} & Top-1 & Top-5\\
\midrule[1.15pt]
\xmark & \cmark & \cmark & \cmark & 70.61 & 92.60 \\ 
\cmark & \xmark & \cmark & \cmark & 70.78 & \textbf{92.92}\\ 
\cmark & \cmark & \xmark & \cmark & 70.63 & 92.80\\ 
\cmark & \cmark & \cmark & \xmark & 70.70 & 92.77\\ 
\cellcolor{grey}\cmark & \cellcolor{grey}\cmark & \cellcolor{grey}\cmark & \cellcolor{grey}\cmark & \cellcolor{grey}\textbf{70.97} & \cellcolor{grey}92.75 \\
\end{tabular}
\vspace{-3mm}
\end{table}

\noindent
\textbf{Pre-training Dataset.} 
In Table~\ref{tab:re_w_pt_ds}(b), we compare randomly initialized encoder and self-supervised MAE pre-trained models on three different datasets, \textit{i.e.,} ImageneNet-1K~\cite{deng2009imagenet}, Kinetics-400~\cite{kay2017k400} and Something-Something v2~\cite{goyal2017something}.
When using the model pre-trained on ImageneNet-1K, the 2D patch embedding layer is inflated to the 3D embedding layer following~\cite{tong2022videomaenju,carreira2017i3d}.
We see that the encoders pre-trained on the video datasets outperform the training from scratch as well as the image-based pre-trained ones. 
Further, the model pre-trained on Something-Something v2 shows better accuracy than the Kinetics-400 pre-trained one, suggesting that the domain gap between target and pre-training datasets found by VideoMAE~\cite{tong2022videomaenju} still exists in our MAR.

\begin{figure}[t]
    \centering
    \includegraphics[width=0.9\linewidth]{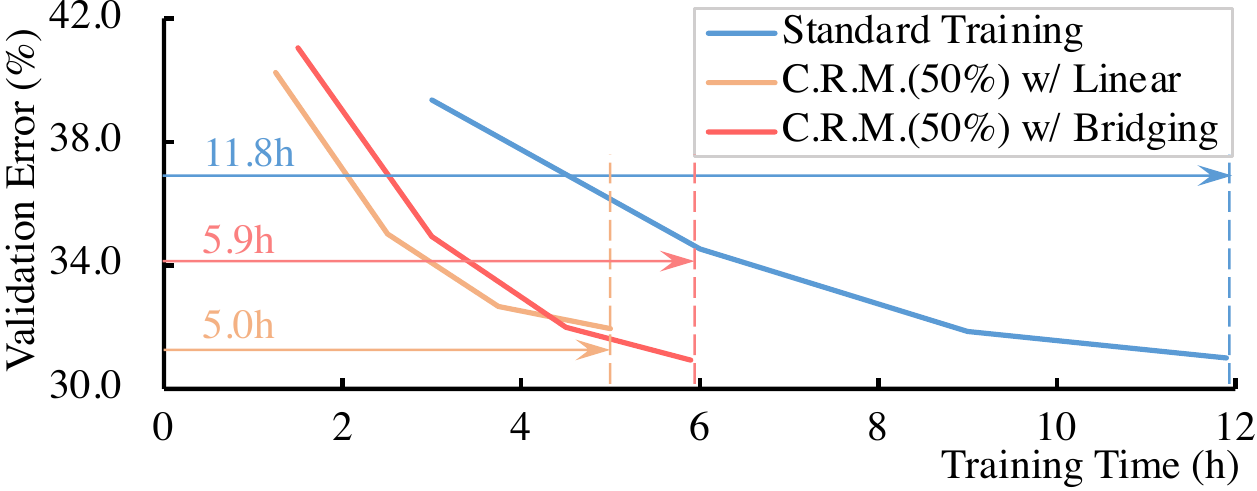}
    \caption{The training time (32 V100 GPUs) vs. validation error rate on Something-Something v2. ``C.R.M.(50\%)'' means the cell running masking with a mask ratio of 50\%. ``Linear'' and ``Bridging'' refer to the linear classifier and bridging classifier, respectively.}
    \label{fig:training_time}
    \vspace{-3mm}
\end{figure}

\noindent
\textbf{Data augmentations.} Although the self-supervised MAE pre-trainings~\cite{he2021mae,tong2022videomaenju,feichtenhofer2022videomaefb} require only multi-scale cropping for training, data augmentation is still one of the crucial factors for transformer models. 
If not specified, our used data augmentations follow the setting in VideoMAE~\cite{tong2022videomaenju}.
In Table 7, we disable four different data augmentations, \textit{i.e.,} RandAugment~\cite{cubuk2020randaugment}, Random Erasing~\cite{zhong2020randomerasing}, MixUp~\cite{zhang2017mixup} and CutMix~\cite{yun2019cutmix} to observe the sensitivity of MAR to data augmentation. 
We observe that each data augmentation can bring a performance gain of 0.2-0.3\%.
In fact, using masks to delete a proportion of the spatio-temporal patches can also be considered one of the data augmentations. 
However, since the masked patches can be easily reconstructed, other data augmentation strategies are still needed.



\begin{table*}[ht]
\caption{\textbf{System-level comparisons on Kinetics-400 action classification.} ``$\rho$'' is the mask ratio. ``FLOPs$\times$Cr.$\times$Cl.'' refers to ``FLOPs$\times$Crops$\times$Clips''. To avoid confusion, we mark the MAE in the video domain as ``MAE-v''.\label{tab:sota_k400}}
\centering
\tablestyle{4.0pt}{0.65}
\begin{tabular}{ccccccccc}
Method &\makecell{Pre-training \\Dataset} & \makecell{Supervised \\ Pre-training}  & Architecture & Input Size & \footnotesize \makecell{FLOPs$\times$Cr.$\times$Cl.\\(G)} & \makecell{Param\\(M)} & \makecell{Top-1\\(\%)} & \makecell{Top-5\\(\%)}\\
\midrule[1.15pt]
NonLocal I3D~\cite{wang2018nonlocal} & ImageNet-1K & \cmark & ResNet101 & $128\times224^2$ & $234\times3\times10$ & 62 & 77.3 & 93.3 \\
TAdaConvNeXt-T~\cite{huang2021tada} & ImageNet-1K & \cmark & ConvNeXt-T & $32\times224^2$ & $94\times3\times4$ & 38 & 79.1 & 93.7 \\
Motionformer~\cite{patrick2021motionformer} & ImageNet-21K & \cmark & ViT-L & $32\times224^2$ & $1185\times3\times10$ & 382 & 80.2 & 94.8 \\
Video Swin~\cite{liu2021videoswin} & ImageNet-1K & \cmark & Swin-B & $32\times224^2$ & $282\times3\times4$ & 88 & 80.6 & 94.6 \\
TimeSformer~\cite{bertasius2021timesformer} & ImageNet-21K & \cmark & ViT-L & $96\times224^2$ & $8353\times3\times1$ & 430 & 80.7 & 94.7 \\
ViViT FE~\cite{arnab2021vivit} & ImageNet-21K & \cmark & ViT-L & $128\times224^2$ & $3980\times3\times1$ & N/A & 81.7 & 93.8 \\
Video Swin~\cite{liu2021videoswin} & ImageNet-21K & \cmark & Swin-L & $32\times224^2$ & $604\times3\times4$ & 197 & 83.1 & 95.9 \\
ip-CSN~\cite{tran2019csn} & - & \xmark & ResNet152 & $32\times224^2$ & $109\times3\times10$ & 33 & 77.8 & 92.8 \\
SlowFast~\cite{feichtenhofer2019slowfast} & - & \xmark & R101+NL & $(16+64)\times224^2$ & $359\times3\times10$ & 60 & 79.8 & 93.9 \\
BEVT~\cite{wang2022bevt} & \scriptsize  IN-1K+DALLE & \xmark & Swin-B & $32\times224^2$ & $282\times3\times5$ & 88 & 80.6 & N/A \\
MViT~\cite{fan2021mvit} & - & \xmark & MViT-B & $32\times224^2$ & $170\times1\times5$ & 37 & 80.2 & 94.4 \\
MViT~\cite{fan2021mvit} & - & \xmark & MViT-B & $64\times224^2$ & $455\times1\times5$ & 37 & 81.2 & 95.1 \\
MaskFeat~\cite{wei2022maskedfeat} & Kinetics-400& \xmark& MViT-L & $16\times224^2$ & $377\times1\times10$ & 218 & 84.3 & 96.3 \\
MaskFeat~\cite{wei2022maskedfeat} & Kinetics-600& \xmark& MViT-L & $16\times224^2$ & $377\times1\times10$ & 218 & 85.1 & 96.6 \\
\midrule
VideoMAE~\cite{tong2022videomaenju} & Kinetics-400 & \xmark & ViT-B & $16\times224^2$ & $180\times3\times5$ & 87 & 80.7 & 94.7 \\
MAE-v~\cite{feichtenhofer2022videomaefb} & Kinetics-400& \xmark& ViT-B & $16\times224^2$ & $180\times3\times7$ & 87 & 81.3 & 94.9 \\
\hdashline
\cellcolor{lightgrey}\textbf{MAR$_{\rho=75\%}$} &\cellcolor{lightgrey}Kinetics-400&\cellcolor{lightgrey}\xmark &\cellcolor{lightgrey}ViT-B &\cellcolor{lightgrey}$16\times224^2$ &\cellcolor{lightgrey}$41\times3\times5$ &\cellcolor{lightgrey}94 &\cellcolor{lightgrey}79.4 &\cellcolor{lightgrey}93.7 \\
\cellcolor{lightgrey}\textbf{MAR$_{\rho=50\%}$} &\cellcolor{lightgrey}Kinetics-400&\cellcolor{lightgrey}\xmark &\cellcolor{lightgrey}ViT-B &\cellcolor{lightgrey}$16\times224^2$ &\cellcolor{lightgrey}$86\times3\times5$ &\cellcolor{lightgrey}94 &\cellcolor{lightgrey}81.0 &\cellcolor{lightgrey}94.4 \\
\midrule
VideoMAE~\cite{tong2022videomaenju} &Kinetics-400& \xmark & ViT-L & $16\times224^2$ & $597\times3\times5$ & 305 & 83.9 & 96.3 \\
MAE-v~\cite{feichtenhofer2022videomaefb} & Kinetics-400& \xmark& ViT-L & $16\times224^2$ & $598\times3\times7$ & 304 & 84.8 & 96.2 \\
\hdashline
\cellcolor{lightgrey}\textbf{MAR$_{\rho=75\%}$} &\cellcolor{lightgrey}Kinetics-400&\cellcolor{lightgrey}\xmark &\cellcolor{lightgrey}ViT-L &\cellcolor{lightgrey}$16\times224^2$ &\cellcolor{lightgrey}$131\times3\times5$ &\cellcolor{lightgrey}311 &\cellcolor{lightgrey}83.9 &\cellcolor{lightgrey}96.0 \\
\cellcolor{lightgrey}\textbf{MAR$_{\rho=50\%}$} &\cellcolor{lightgrey}Kinetics-400&\cellcolor{lightgrey}\xmark &\cellcolor{lightgrey}ViT-L &\cellcolor{lightgrey}$16\times224^2$ &\cellcolor{lightgrey}$276\times3\times5$ &\cellcolor{lightgrey}311 &\cellcolor{lightgrey}\textbf{85.3} &\cellcolor{lightgrey}\textbf{96.3} \\
\midrule
MAE-v~\cite{feichtenhofer2022videomaefb} & Kinetics-400& \xmark& ViT-H & $16\times224^2$ & $1193\times3\times7$ & 632 & 85.1 & 96.6 \\

\end{tabular}
\end{table*}

\begin{table*}[ht]
\caption{\textbf{System-level comparisons on Something-Something v2 (SSv2).} ``$\rho$'' is the mask ratio. ``FLOPs$\times$Cr.$\times$Cl.'' refers to ``FLOPs$\times$Crops$\times$Clips''. To avoid confusion, we mark the MAE in the video domain as ``MAE-v''. \label{tab:sota_ssv2}}
\centering
\tablestyle{4.0pt}{0.65}
\begin{tabular}{ccccccccc}
Method & \makecell{Pre-training \\Dataset} & \makecell{Supervised \\ Pre-training} & Architecture & Input Size & \footnotesize \makecell{FLOPs$\times$Cr.$\times$Cl.\\(G)} & \makecell{Param\\(M)} & \makecell{Top-1\\(\%)} & \makecell{Top-5\\(\%)}\\
\midrule[1.15pt]
TimeSformer~\cite{bertasius2021timesformer}&ImageNet-21K & \cmark & ViT-L & $64\times224^2$ & $5549\times3\times1$ & 430 & 62.4 & N/A \\
SlowFast~\cite{feichtenhofer2019slowfast} & Kinetics-400 & \cmark & ResNet101 & $(8+32)\times224^2$ & $106\times3\times1$ & 53 & 63.1 & 87.6 \\
TAdaConvNeXt-T~\cite{huang2021tada} & ImageNet-1K& \cmark & ConvNeXt-T & $32\times224^2$ & $94\times3\times2$ & 38 & 67.1 & 90.4 \\
Motionformer~\cite{patrick2021motionformer}&\scriptsize  IN-21K+K400&\cmark&ViT-L&$32\times224^2$&$1185\times3\times1$&382 & 68.1 & 91.2 \\
MViT~\cite{fan2021mvit} & Kinetics-600 &\cmark & MViT-B-24&$32\times224^2$ & $236\times3\times1$&53&68.7&91.5\\
Video Swin~\cite{liu2021videoswin} &\scriptsize  IN-21K+K400& \cmark & Swin-B & $32\times224^2$ & $321\times3\times1$ & 88 & 69.6 & 92.7 \\
BEVT~\cite{wang2022bevt} &\scriptsize IN-1K+K400+DALLE& \xmark& Swin-B & $32\times224^2$ & $321\times3\times1$ & 88 & 70.6 & N/A \\
MaskFeat~\cite{wei2022maskedfeat} & Kinetics-400& \xmark& MViT-L & $40\times312^2$ & $2828\times3\times1$ & 218 & 74.4 & 94.6 \\
\midrule
VideoMAE~\cite{tong2022videomaenju} &SSv2& \xmark & ViT-B & $16\times224^2$ & $180\times3\times2$ & 87 & 70.3 & 92.7 \\
\hdashline
\cellcolor{lightgrey}\textbf{MAR$_{\rho=75\%}$} & \cellcolor{lightgrey}SSv2& \cellcolor{lightgrey}\xmark & \cellcolor{lightgrey}ViT-B & \cellcolor{lightgrey}$16\times224^2$ & \cellcolor{lightgrey}$41\times3\times2$ & \cellcolor{lightgrey}94 & \cellcolor{lightgrey}69.5 & \cellcolor{lightgrey}91.9 \\
\cellcolor{lightgrey}\textbf{MAR$_{\rho=50\%}$} & \cellcolor{lightgrey}SSv2& \cellcolor{lightgrey}\xmark & \cellcolor{lightgrey}ViT-B & \cellcolor{lightgrey}$16\times224^2$ & \cellcolor{lightgrey}$86\times3\times2$ & \cellcolor{lightgrey}94 & \cellcolor{lightgrey}\textbf{71.0} & \cellcolor{lightgrey}\textbf{92.8} \\
\midrule
VIMPAC~\cite{tan2021vimpac} &\scriptsize HowTo100M+DALLE& \xmark& ViT-L & $10\times224^2$ & $\text{N/A}\times3\times10$ & 307 & 68.1 & N/A \\
MAE-v~\cite{feichtenhofer2022videomaefb} & Kinetics-400& \xmark & ViT-L & $16\times224^2$ & $598\times3\times1$ & 304 & 72.1 & 93.9 \\
VideoMAE~\cite{tong2022videomaenju} & SSv2& \xmark & ViT-L & $16\times224^2$ & $597\times3\times2$ & 305 & 74.2 & 94.7 \\
\hdashline
\cellcolor{lightgrey}\textbf{MAR$_{\rho=75\%}$} & \cellcolor{lightgrey}Kinetics-400& \cellcolor{lightgrey}\xmark & \cellcolor{lightgrey}ViT-L & \cellcolor{lightgrey}$16\times224^2$ & \cellcolor{lightgrey}$131\times3\times2$ & \cellcolor{lightgrey}311 & \cellcolor{lightgrey}73.8 &\cellcolor{lightgrey}94.4  \\
\cellcolor{lightgrey}\textbf{MAR$_{\rho=50\%}$} & \cellcolor{lightgrey}Kinetics-400& \cellcolor{lightgrey}\xmark & \cellcolor{lightgrey}ViT-L & \cellcolor{lightgrey}$16\times224^2$ & \cellcolor{lightgrey}$276\times3\times2$ & \cellcolor{lightgrey}311 & \cellcolor{lightgrey}\textbf{74.7} & \cellcolor{lightgrey}\textbf{94.9} \\
\midrule
MAE-v~\cite{feichtenhofer2022videomaefb} & Kinetics-400& \xmark & ViT-H & $16\times224^2$ & $1193\times3\times1$ & 632 & 74.1 & 94.5 \\
\end{tabular}
\vspace{-3mm}
\end{table*}

\begin{table}[!t]
    \caption{\textbf{Comparisons to other state-of-the-art methods on UCF101 and HMDB51.}}
    \label{tab:ucf_hmdb}
    \centering
    \tablestyle{4pt}{1.0}
    \begin{tabular}{ccccc}
       Method  & Backbone & Extra Data & UCF101 & HMDB51 \\
       \midrule[1.15pt]
       CoCLR~\cite{coclr} & S3D-G & UCF101 & 81.4 & 52.1 \\
       Vi$^2$CLR~\cite{diba2021vi2clr} & S3D & UCF101 & 82.8 & 52.9 \\
       MoSI~\cite{huang2021mosi} & R(2+1)D & - & 82.8 & 51.8 \\
       \midrule
       VideoMAE~\cite{tong2022videomaenju}  & ViT-B & - & 90.8 & 61.1\\ 
       \hdashline
       \cellcolor{lightgrey}\textbf{MAR}$_{\rho=50\%}$  & \cellcolor{lightgrey}ViT-B & \cellcolor{lightgrey}- & \cellcolor{lightgrey}\textbf{91.0} & \cellcolor{lightgrey}\textbf{61.4}\\ 
       \midrule
       \midrule
       MemDPC~\cite{han2020memdpc} & R-2D3D&K400&86.1&54.5\\
       CoCLR~\cite{coclr} & S3D-G & K400 & 87.9 & 54.6 \\
       Vi$^2$CLR~\cite{diba2021vi2clr} & S3D & K400 & 89.1 & 55.7 \\
       ParamCrop~\cite{qing2021paramcrop} & S3D-G & K400 & 91.3 & 63.4 \\
       RSPNet~\cite{chen2020rspnet} & S3D-G & K400 & 93.7 & 64.7\\
       HiCo~\cite{qing2022hico} & S3D-G & UK400 & 91.0 & 66.5\\
       CVRL~\cite{qian2021cvrl} & Slow-R152 & K600 & 94.4 & 70.6\\
       $\rho$BYOL~\cite{feichtenhofer2021largescale} & Slow-R50 & K400 & 94.2 & 72.1 \\
       \midrule
       VideoMAE~\cite{tong2022videomaenju}  & ViT-B & K400 & \textbf{96.1} & 73.3\\ 
       \hdashline
       \cellcolor{lightgrey}\textbf{MAR}$_{\rho=50\%}$  & \cellcolor{lightgrey}ViT-B & \cellcolor{lightgrey}K400 & \cellcolor{lightgrey}95.9 & \cellcolor{lightgrey}\textbf{74.1}\\ 
    \end{tabular}
\end{table}

\noindent
\textbf{Wall-clock training time.} Figure~\ref{fig:training_time} compares the training time consumed by the different training methods on Something-Something v2 dataset.
It can be observed that MAR can \textit{reduce} half of the training cost.
Specifically, the standard training takes 11.8 hours, while our MAR only needs 5.9 hours and achieves better accuracy than the standard scheme.
Although the introduction of the bridging classifier increases the training time slightly, it is still acceptable compared to the standard training scheme.

\noindent
\textbf{Visualizations.} We qualitatively visualize the reconstructed videos by reconstruction branch in Figure~\ref{fig:vis_reconstruction}.
We observe that even if the cell running masking discards 50\% of the spatio-temporal patches, the reconstructed videos can still fully express the high-level semantics of the videos. 
This suggests that it is not necessary for the encoder to operate all spatio-temporal patches.

\subsection{Comparisons with the Previous Methods}
Table~\ref{tab:sota_k400} and Table~\ref{tab:sota_ssv2} compare our training scheme with other state-of-the-art methods on Kinetics-400~\cite{kay2017k400} and Something-Something v2~\cite{goyal2017something}. 
The relevant settings are listed in detail for comparison, including network architectures and calculation costs.
We can draw the following observations from the table.
First, with the same encoders, \textit{i.e.,} pre-trained by VideoMAE, MAR can consistently outperform the standard training scheme by at least 0.3\% with only around 47\% of GFLOPs.
Second, MAR can easily scale up to large models and achieve more improvements. 
The ViT-Large trained by MAR on Kinetics-400 dataset surpasses standard training by 1.4\%.
With a large mask ratio (\textit{i.e.,} 75\%), our ViT-Large has an advantage over the standard trained ViT-Base of 2.6\% (81.3\% \textit{vs.} 83.9\%) on Kinetics-400, while our ViT-Large saves 27\% of computation overhead (180 GFLOPs \textit{vs.} 131 GFLOPs).
Moreover, it is also worth noting that our trained ViT-Large models even exceed ViT-Huge by 0.2\% on Kinetics-400 and Something-Something v2.
Especially, our ViT-Large only costs 14\% GFLOPs compared to ViT-Huge with  standard training.
Third, the models trained by MAR achieve superior performance on both datasets compared to previous approaches under the similar GFLOPs, even though they use the supervised pre-training on larger datasets.
In addition, comparisons on two small video datasets, \textit{i.e.,} UCF101~\cite{soomro2012ucf101} and HMDB51~\cite{kuehne2011hmdb} are also reported in Table~\ref{tab:ucf_hmdb}.
It can be observed that our MAR still leads to on par or better performance under multiple settings on these two small datasets.
We speculate that the small datasets with limited videos can be easily reconstructed by models, thus leading to overfitting, especially for UCF101 datasets with simple backgrounds.
Therefore, our MAR is superior on large-scale datasets.

\section{Conclusion}
In this work, we propose Masked Action Recognition (MAR), a simple and computationally friendly training scheme for vanilla Vision Transformers (ViTs) in videos.
MAR is investigated from two perspectives of ViTs, \textit{i.e.,} reducing the number of input patches and bridging the semantic gaps of output features.
For the former, the cell running masking strategy is designed to generate spatio-temporal interleaved masks, which preserves the spatio-temporal correlations in videos. 
For the latter, the lightweight bridging classifier is proposed to bridge the semantic gaps between encoded features and specialized classification features. 
Empirical results show that MAR costs only 47\% of the computation and exceeds the performance of the standard training scheme. 
In addition, strong generalizations of MAR have also been demonstrated on several video datasets with different scales.
Overall, this work exploits the powerful context modelling capability of ViTs, and significantly improves the training and testing efficiency with better performance.
In future works, to further save computational costs with a less performance penalty, semantic-based mask ratios and masking maps are worthwhile prospects to be explored.


\bibliographystyle{IEEEtran}
\bibliography{egbib}

\end{document}